\pdfoutput=1

\documentclass[11pt]{article}

\usepackage[]{acl}

\usepackage{times}
\usepackage{latexsym}

\usepackage[T1]{fontenc}

\usepackage[utf8]{inputenc}

\usepackage{microtype}

\usepackage{tabularx}
\usepackage{graphicx}
\usepackage{array}
\usepackage{float}
\restylefloat{table}
\usepackage{adjustbox}
\usepackage{array, multirow}
\usepackage{comment}
\usepackage{xcolor}
\usepackage{amsmath}
\usepackage{bm}
\DeclareMathOperator{\argsort}{argsort}
\usepackage[english]{babel}
\usepackage{blindtext}
\usepackage{bbm}

\usepackage{color}
\usepackage{tcolorbox}
\usepackage{CJK}
\usepackage{adjustbox}
\usepackage{booktabs}
\tcbset{width=0.4\textwidth,boxrule=0pt,colback=red,arc=0pt,auto outer arc,left=0pt,right=0pt,boxsep=5pt}
\usepackage{cleveref}

\usepackage{microtype}

\title{Dynamically Refined Regularization for Improving \\ Cross-corpora Hate Speech Detection}

  \author{Tulika Bose \textsuperscript{$\dagger$} \quad Nikolaos Aletras \textsuperscript{$\ddagger$} \quad Irina Illina \textsuperscript{$\dagger$} \quad Dominique Fohr \textsuperscript{$\dagger$}  \\
  \textsuperscript{$\dagger$} Universite de Lorraine, CNRS, Inria, LORIA, F-54000 Nancy, France\\
  \textsuperscript{$\ddagger$}University of Sheffield, United Kingdom\\
  {\tt \{tulika.bose, illina, dominique.fohr\}@loria.fr}\\ 
  {\tt n.aletras@sheffield.ac.uk}
}

\date{}

\begin{document}
\maketitle
\begin{abstract}

\emph{{\bf Warning: } this paper contains content that may be offensive and distressing.}

Hate speech classifiers exhibit substantial performance degradation when evaluated on datasets different from the source. This is due to learning spurious correlations between words that are not necessarily relevant to hateful language, and hate speech labels from the training corpus. Previous work has attempted to mitigate this problem by regularizing 
specific terms
from pre-defined static dictionaries. While this has 
been demonstrated to 
improve the generalizability of classifiers, the coverage of such methods is limited and the dictionaries require regular manual updates from human experts. In this paper, we propose to automatically identify and reduce spurious correlations using attribution methods with dynamic refinement of the list of terms that need to be regularized during training. Our approach is flexible and improves the cross-corpora performance over previous work independently and in combination with pre-defined 
dictionaries.\footnote{Code is available here: \url{https://github.com/tbose20/D-Ref}} 

\end{abstract}


\section{Introduction}
\label{intro}

The relative sparsity of hateful content in the real world requires crawling of many of the standard hate speech corpora through keyword-based sampling \citep{Poletto2021ResourcesAB}, rather than random sampling. Thus, hate speech classifiers \citep{ dsa-etal-2020-towards, mozafari2019bert, 10.1145/3041021.3054223} often learn spurious correlations from the training corpus \citep{wiegand-etal-2019-detection} leading to a substantial performance degradation when evaluated on a corpus with a different distribution \citep{yin2021generalisable, bose-etal-2021-unsupervised, app10124180, 10.1145/3331184.3331262,swamy-etal-2019-studying, karan-snajder-2018-cross}. 

Recent work has proposed regularization mechanisms to penalize spurious correlations by attempting to explain model predictions using feature attribution methods \citep{ijcai2017-371, pmlr-v119-rieger20a, NEURIPS2020_075b051e}. These methods assign importance scores to input tokens that contribute more towards a particular prediction~\citep{NIPS2017_8a20a862}. 
For instance, \citet{liu-avci-2019-incorporating} 
penalize the attributions assigned to tokens contained in a manually curated dictionary consisting of group identifiers (e.g. women, jews) that are often known to be targets of hate. \citet{kennedy-etal-2020-contextualizing} extract group identifiers manually from the top tokens indicated by a bag-of-words logistic regression model trained on the source corpus.
However, regularizing only group identifiers
limits the coverage of such approaches, and may not capture other forms of corpus-specific correlations learned by the classifier limiting its performance on a new corpus. Moreover, such manually curated lists may not always remain up-to-date because new terms emerge frequently~\citep{Grieve2018MappingLI}. While \citet{yao2021refining} do not use such lists for refining models in different target-domains, their method still requires input from human annotators.

In this paper, we hypothesize that the classification errors 
in a small annotated subset 
from the target can reveal spurious correlations between tokens and hate speech labels learned from the source (see Table \ref{spu-examples}). 
\begin{table}[!t]
\small
\centering
\begin{tabularx}{\columnwidth}{p{3.7cm} c c}
\hline \textbf{Target corpus utterances}& \textbf{Actual} & \multicolumn{1}{p{1.5cm}}{\textbf{Predicted}} \\ 
\colorbox{purple!46.25}{Genocide} \colorbox{purple!4.1000000000000005}{is} \colorbox{purple!1.95}{never} \colorbox{purple!27.150000000000002}{ok}  & non-hate & hate \\
\colorbox{purple!30.020000000000003}{Women} \colorbox{purple!9.08}{are} \colorbox{purple!5.680000000000001}{goddesses} & non-hate & hate \\
\hline
\end{tabularx}
\caption{\label{spu-examples} 
Spurious correlations learned by the source classifier between the shaded tokens and the hate label. } 
\end{table}
    To this end, we propose Dynamic Model Refinement (D-Ref), a new method to identify and penalize spurious tokens using feature attribution methods.
    We demonstrate that D-Ref improves the overall cross-corpora performance independently and in combination with pre-defined dictionaries.

\section{Dynamic Model Refinement (D-Ref)}

In this section, we describe the general theoretical framework of the proposed approach. We assume that during training our hate speech classification model has access to the source training corpus $D_S^{train}$ and a small validation set $D_T^{val}$ from a target corpus with different distribution, following a similar setting to \citet{maharana-bansal-2020-adversarial}. 
Our Dynamic Model Refinement (D-Ref) approach consists of 2 recurring steps across epochs: (i) we first extract a set of spurious tokens 
using $D_T^{val}$
at the end of every epoch; 
and (ii) then we penalize the extracted tokens
during the next epoch.

\subsection{Extraction of Spurious Tokens} 

\textbf{Global token-ranking in source corpus:} 
We first begin with identifying the tokens from $D_S^{train}$ that are highly correlated with hate/non-hate labels. These tokens are suitable candidates for causing source-specific spurious correlations, restricting generalizability to a new corpus. 

For that purpose, at the end of every training epoch $ep_i$, we first obtain the global class-specific ranked list of tokens from $D_S^{train}$. This is achieved by computing global attributions per token $tok$ and class $c$ ($\text{gl}\mbox{-}\text{atr}_{tok}^{c}$)
from its attribution per instance $j$ ($\text{loc\mbox{-}atr}_{tok}^j$)  averaged across all training instances classified as $c$ by the source model trained until $ep_i$:
\begin{equation}\label{eq1}
  \resizebox{0.85\columnwidth}{!}{%
  $\text{gl\mbox{-}atr}_{tok}^c = \frac{\sum_{j=1}^{|D_S^{train}|}\mathbbm{1}_{\hat{y_j}=c}\text{loc\mbox{-}atr}_{tok}^j \forall  \text{occurrence of $tok$ in j}}{\sum_{j=1}^{|D_S^{train}|}\mathbbm{1}_{\hat{y_j} = c} \#(\text{occurrence of $tok$ in j})}$%
  }
\end{equation}
Here $c$ $\in$ \{hate, non-hate\}, $\hat{y}$ is the predicted class and $\mathbbm{1}$ is the indicator function.
Prior to this, $\text{loc\mbox{-}atr}_{tok}^j$ are individually normalized using sigmoid to obtain values in a closed range. Rarely occurring tokens
and stop-words are not considered for the global ranking. The $\text{gl\mbox{-}atr}_{tok}^c$ values are sorted from the highest globally attributed token to the lowest, which yields two ranked token-lists $[{gl\mbox{-}hate}, {gl\mbox{-}nhate}]_{ep_i}$. 

\paragraph{Instance-level local ranking in target corpus:}
We hypothesize that tokens highly correlated with hate/non-hate classes in the source, but also causing mis-classifications in the target, should most likely 
contribute to spurious source-specific correlations, and may not be important for hate speech labels. Thus, we identify the tokens that cause mis-classifications in $D_T^{val}$, and then obtain a list of spurious tokens \textit{dynamically} 
after every epoch $ep_i$.

We rank the tokens in the target instances from $D_T^{val}$ based on their
$\text{loc\mbox{-}atr}_{tok}^j$,
starting from the highest attributed token per instance $j$ to the lowest.
The top $k$ tokens in $j$ is given by $tok_{top_k}^j = top_k[\argsort(\text{loc\mbox{-}atr}_{tok}^j)]$, 
where $k$ is a hyper-parameter 
in $D_T^{val}$. 
We treat the two error cases of False Positives (FP) and False Negatives (FN) separately. Here the hate class is considered as the positive class. 

Since the tokens responsible for FP may also be important for the True Positives (TP), we only extract those 
that have high attributions for FP, but not for TP.
Further, another filtering step is applied, where only the tokens common to the top $N$ from the ranked $gl\mbox{-}hate$ are extracted.
This results in discarding the tokens that 
may not be globally correlated with a class with respect to the source model.  So
$tok_{FP} = [tok \in tok_{top_k}^{j_{FP}} \And tok \not\in tok_{top_k}^{j_{TP}}] \cap top_N(gl\mbox{-}hate)$ $\forall$ instances $j$ in $D_T^{val}$.
Similarly, top $k$ tokens corresponding to FN instances are extracted, wherein those common to TN are discarded, and subsequent filtering based on the $gl\mbox{-}nhate$ is performed, i.e. $tok_{FN} = [tok \in tok_{top_k}^{j_{FN}} \And tok \not\in tok_{top_k}^{j_{TN}}] \cap top_N(gl\mbox{-}nhate)$ $\forall$ $j$.
This step thus yields a list of 
possible spurious tokens at the end of $ep_i$,  $S_{ep_i} = [tok_{FP},tok_{FN}]_{ep_i}$.


\subsection{Penalizing the Extracted Spurious Tokens} 
In this step, we attempt to reduce the importance assigned, by the source model, to the extracted spurious tokens by penalizing the terms in $S_{ep_i}$
during the next epoch $ep_{i+1}$. We propose three different ways for token penalization: 

\paragraph{Tok-mask:} 
In this case, we simply mask the tokens from  $S_{ep_i}$ present in $D_S^{train}$  after every $ep_i$ and then train the source model during $ep_{i+1}$. 

\paragraph{Reg:} 
Since token masking might eliminate substantial information, 
we regularize the model using $S_{ep_i}$. The attributions assigned to these terms
are pushed towards zero by the following
learning objective on $D_S^{train}$:
\begin{equation}\label{eq2}
\resizebox{0.85\hsize}{!}{%
    $\mathcal{L} = \mathcal{L^{'}} + \lambda \mathcal{L}_{\textrm{atr}}\left(\text{t}\right); \text{t} \in S_{\text{ep}_\text{i}}; \mathcal{L}_\textrm{atr} = \sum \limits_{\text{t}\in S_{\text{ep}_\text{i}}} \phi \left(\text{t}\right)^\textsuperscript{2}$%
    }
\end{equation}
where $\mathcal{L^{'}}$ is the 
classification loss and $\mathcal{L_{\text{atr}}}$
is the attribution loss. Here $\phi \left(\text{t}\right)$ is the attribution score 
for the token t. Intuitively, 
this should reduce the importance of tokens contributing to source-specific patterns 
and encourage learning more general information. Both losses are computed over $D_S^{train}$.

\paragraph{Comb:}  
We finally combine $S_{ep_i}$ with the pre-defined group identifiers from \citet{liu-avci-2019-incorporating} and \citet{kennedy-etal-2020-contextualizing} to perform regularization using Equation \ref{eq2}.

We surmise that repeating these steps at the end of every epoch should reduce the source-specific correlations while the source model gets trained. We use three different attribution methods: 
\paragraph {(i) \textbf{Scaled Attention ($\alpha \nabla \alpha$})} \citep{serrano-smith-2019-attention}:
Here attention weights $\alpha_i$ are scaled with their corresponding gradients $\nabla\alpha_i = \frac{\delta \hat{y}}{\delta \alpha_i}$, where $\hat{y}$ is the predicted label. \citet{serrano-smith-2019-attention} show that combining an attention weight with
its gradient can better indicate token importance for
model predictions, compared to only using the attention weights. \paragraph{(ii) \textbf{Integrated Gradients (IG)}} \citep{pmlr-v70-sundararajan17a}: 
This method is based on the notion that the gradient of a prediction function with respect to input can indicate the sensitivity of the prediction for each input dimension. As such, it aggregates the gradients along a path from an
uninformative 
reference input (e.g. zero embedding vector) towards the 
actual input such that the predictions change from uncertainty to certainty.  
\paragraph{(iii) \textbf{Deep Learning Important
FeaTures (DeepLIFT/DL)}} \citep{pmlr-v70-shrikumar17a}: This aims to explain the difference in the output from a reference output in terms of the difference of the input and a reference input. Given a target output neuron $t$, a reference activation $t^0$ of $t$, and $\Delta{t} = t-t^0$, it computes the contribution scores $C_{\Delta{x_i}\Delta{t}}$ of each input neuron \textit{$x_i$} that are necessary and sufficient to compute \textit{t}, such that 
$\sum_{i=1}^n C_{\Delta{x_i}\Delta{t}} = \Delta{t}.
$
The 
reference input could be the zero embedding vector.

\section{Experiments and Results}

\subsection{Experimental Setup}
\label{exp_setup}

\paragraph{Data}
\label{corp}
We use three standard hate speech corpora: \textit{HatEval} \citep{basile-etal-2019-semeval}, \textit{Waseem} \citep{waseem} and \textit{Dynamic} \citep{vidgen-etal-2021-learning}. Following previous work by \citet{wiegand-etal-2019-detection, swamy-etal-2019-studying}, we consider the detection of hate vs non-hate, where the hate class covers all forms of hate. We split \textit{Waseem} (26.8\% hate) into train (80\%; 8720), val (10\%; 1090) and test (10\%; 1090) sets as no standard splits are provided. We use the original splits for \textit{HatEval} (42.1\% hate; train: 8993\footnote{We remove the instances that contain only URLs, reducing the train instances from 9000 to 8993.}, val: 1000; test: 3000) and \textit{Dynamic} (54.4\% hate; train: 32497, val: 1016, test: 4062). We reduce the size of available $D_T^{val}$ in \textit{Dynamic} by randomly sampling 25\% of the validation set (4064). 
We remove URLs, split hashtags into words using the CrazyTokenizer\footnote{\url{https://redditscore.readthedocs.io}}, remove infrequent Twitter handles, punctuation marks and numbers, and convert text into lower-case. 
See Appendix \ref{data_desc} for a detailed discussion on the corpora.



\paragraph{Baselines}
We compare D-Ref with the following baselines: \textbf{(i) BERT   Van-FT} \citep{devlin-etal-2019-bert}: vanilla fine-tuning on $D_S^{train}$ without regularization; \textbf{(ii)} Convolutional Neural Network with regularization of pre-defined group identifier terms using IG for feature attribution \citep{liu-avci-2019-incorporating}; \textbf{(iii)} BERT using two variations for regularization: (a) all the mentioned group identifiers, (b) group identifiers extracted from the top features of a bag-of-words logistic regression trained on each individual corpus~\cite{kennedy-etal-2020-contextualizing}\footnote{We use Sampling and Occlusion \citep{Jin2020Towards}.}; \textbf{(iv) $\bm{\chi^2}$-test} with one degree of freedom and Yate's correction \citep{jbp:/content/journals/10.1075/ijcl.6.1.05kil} to extract tokens $tok$ from $D_S^{train}$ that reject the null hypothesis with 95\% confidence. The null hypothesis states that in terms of $tok$, both $D_S^{train}$ and $D_T^{val}$ are random samples of the same larger population. We, then, regularize the attribution scores\footnote{We use DL as it yields comparable or higher overall improvements taking Table \ref{Results_Dyn-Ref} and Table \ref{Results_DA} together.} assigned to these terms, with BERT.  
\textbf{(v) Pre-def}: BERT with regularizing the combined pre-defined group identifiers from (ii) and (iii).

\begin{table*}[!t]
\scriptsize
\centering
\begin{tabularx}{0.95\textwidth}{p{0.6cm}|p{2.5cm}|p{1.4cm}|p{1.4cm}|p{1.4cm}|p{1.4cm}|p{1.4cm}|p{1.4cm}|c}
\toprule
\multicolumn{2}{l|}{\textbf{Approaches}} & \textbf{H \textrightarrow D} & \textbf{D \textrightarrow H} & \textbf{H \textrightarrow W} & \textbf{W \textrightarrow H} & \textbf{D \textrightarrow W}&\textbf{W \textrightarrow D} & \textbf{Average}  \\
\hline
\multicolumn{2}{l|}{BERT Van-FT} &53.2$\pm$1.0& 63.3$\pm$1.8&67.5$\pm$5.1&52.6$\pm$2.4&60.3$\pm$1.0&46.7$\pm$4.0& 57.3\\
\multicolumn{2}{l|}{\citet{liu-avci-2019-incorporating}} & 45.1$\pm$4.5 & 59.5$\pm$0.7 & 57.2$\pm$3.8 & 52.6$\pm$0.8 &57.1$\pm$2.7 & 39.6$\pm$2.0& 51.9\\
\multicolumn{2}{l|}{\citet{kennedy-etal-2020-contextualizing} (a)} & 52.2$\pm$1.2& 62.0$\pm$1.6&62.7$\pm$2.9 & 50.1$\pm$6.8 & 53.5$\pm$2.0& 45.1$\pm$2.3& 54.3\\
\multicolumn{2}{l|}{\citet{kennedy-etal-2020-contextualizing} (b)} & 52.0$\pm$3.8 & 61.9$\pm$1.7&63.6$\pm$3.7 & 54.8$\mbox{*}\pm$1.6 &57.0$\pm$1.7 & 46.8$\pm$1.9 & 56.0\\
\multicolumn{2}{l|}{BERT $\chi^2$-test} & 55.4$\mbox{*}\pm$1.1 & 65.0$\mbox{*}\pm$1.0 & 68.1$\pm$1.3 & 53.7$\pm$2.1 & 60.4$\pm$2.8 & 45.2$\pm$2.8 & 58.0\\
\hline
\hline
\multicolumn{2}{l|}{Pre-def ($\alpha \nabla \alpha$)} & 54.6$\mbox{*}\pm$1.3 & \textbf{65.1}$\mbox{*}\pm$1.1 & 69.6$\pm$3.4 & 54.4$\mbox{*}\pm$1.2 & \textbf{61.9}$\pm$1.6 &47.2$\pm$3.1&58.8\\
\multicolumn{2}{l|}{D-Ref-Tok-mask ($\alpha \nabla \alpha$)} & 53.8$\pm$0.6 & 64.9$\mbox{*}\pm$0.7 & 68.9$\pm$3.3 &53.6$\pm$3.0&59.6$\pm$2.2&45.8$\pm$3.7&57.8\\
\multicolumn{2}{l|}{D-Ref-Reg ($\alpha \nabla \alpha$)} &$54.9\mbox{*}\pm$1.2& \textbf{65.1}$\mbox{*}\pm$0.9&68.6$\pm$4.0&54.1$\mbox{*}\pm$1.0 &60.9$\pm$1.5 &\textbf{48.7}$\mbox{*}\pm$4.3&58.7\\
\multicolumn{2}{l|}{D-Ref-Comb ($\alpha \nabla \alpha$)} &\textbf{55.0}$\mbox{*}\pm$1.6 & 64.7$\mbox{*}\pm$1.2& \textbf{69.9}$\pm$1.6&\textbf{55.3}$\mbox{*}\pm$1.3&61.0$\pm$2.8&48.1$\mbox{*}\pm$1.0 &\textbf{59.0}\\
\hline
\hline
\multicolumn{2}{l|}{Pre-def (IG)} &55.7$\mbox{*}\pm$1.4  & 63.5$\pm$2.8 & \textbf{69.7}$\pm$2.2 & 51.7$\pm$2.7&60.3$\pm$2.2 &44.6$\pm$3.0&57.6\\
\multicolumn{2}{l|}{D-Ref-Tok-mask (IG)} &56.3$\mbox{*}\pm$2.3  & 64.5$\mbox{*}\pm$1.8 &68.3$\pm$2.0&52.3$\pm$2.3 &59.3$\pm$1.3&48.2$\mbox{*}\pm$2.1&58.2\\
\multicolumn{2}{l|}{D-Ref-Reg (IG)} & \textbf{56.4}$\mbox{*}\pm$1.4 & \textbf{65.5}$\mbox{*}\pm$0.8 &69.2$\pm$2.5  &\textbf{53.8}$\mbox{*}\pm$0.7&60.6$\pm$1.7 &47.7$\pm$3.6&58.9\\
\multicolumn{2}{l|}{D-Ref-Comb (IG)}&55.7$\mbox{*}\pm$0.8 & 63.7$\pm$2.4& 69.1$\pm$2.3&52.6$\pm$2.3 &\textbf{61.4}$\pm$2.5&\textbf{51.4}$\mbox{*}\pm$3.6&\textbf{59.0} \\
\hline \hline
\multicolumn{2}{l|}{Pre-def (DL)} & 54.2$\pm$1.6 & 64.0$\pm$1.9 &68.1$\pm$1.5 &52.9$\pm$1.2&62.0$\pm$1.8 &44.5$\pm$1.3&57.6\\
\multicolumn{2}{l|}{D-Ref-Tok-mask (DL)} & 55.1$\mbox{*}\pm$1.4 & \textbf{64.9}$\mbox{*}\pm$1.7 & 67.2$\pm$3.6& 52.1$\pm$1.9 &60.5$\pm$2.5& 47.2$\pm$3.1 & 57.8\\
\multicolumn{2}{l|}{D-Ref-Reg (DL)} & 54.2$\pm$1.6 & 64.8$\mbox{*}\pm$0.8 & \textbf{70.7}$\mbox{*}\pm$2.7 &51.4$\pm$0.7 & \textbf{62.3}$\mbox{*}\pm$2.5 &47.1$\pm$5.5 & 58.4\\
\multicolumn{2}{l|}{D-Ref-Comb (DL)} & \textbf{55.4}$\mbox{*}\pm$1.8  & 64.0$\pm$0.9 & 69.5$\pm$3.3 & \textbf{54.0}$\mbox{*}\pm$0.8 & 61.5$\pm$2.3 &\textbf{48.1}$\mbox{*}\pm$2.7 & \textbf{58.8} \\
\hline
\end{tabularx}
\caption{\label{Results_Dyn-Ref}
Macro-F1 ($\pm$std-dev) on source \textrightarrow target pairs (H : HatEval, D : Dynamic, W : Waseem). \textbf{Bold} denotes the best performing approach in each column for every feature attribution method. \mbox{*} denotes statistical significance compared to Van-FT with paired bootstrap \citep{dror-etal-2018-hitchhikers, efron1994introduction}, 95\% confidence interval.}
\end{table*}

\paragraph{Model training} We use pre-trained BERT \citep{devlin-etal-2019-bert} for our approach.  We train all the models over $D_S^{train}$ from the source and evaluate over $D_T^{test}$ from the target.
The best model 
for all the baselines and D-Ref
are selected by tuning over $D_T^{val}$. 
See Appendix \ref{hyper-param} on hyper-parameter tuning.

\subsection{Cross-corpora Predictive Performance}
\label{results_cross}
Table \ref{Results_Dyn-Ref} presents macro-F1 scores across five random initializations of each experiment 
using six cross-corpora pairs. 
We observe that overall, all feature-attribution methods with D-Ref yield improved performance compared to Van-FT and other baselines. While $\chi^2$ yields improvements over Van-FT,
D-Ref still displays better performance in most of the cases. This could be attributed to the fact that although the terms obtained through the $\chi^2$ test from the source indicate differences across domains,
they may not necessarily be important for the prediction of hate/ non-hate labels by the source model, and may not contribute to source-specific spurious correlations.  

We find that D-Ref-Reg with IG and DL achieves better average macro-F1 of 58.9 and 58.4 respectively, 
compared to the corresponding Pre-def (IG) and Pre-Def (DL) that obtain  an average of 57.6. D-Ref-Reg ($\alpha\nabla\alpha$) provides an average macro-F1 of 58.7, comparable to Pre-def ($\alpha \nabla\alpha$) with 58.8. However, D-Ref-Reg achieves significantly improved scores in more cases, as compared to Pre-def using all the attribution methods, i.e. 4/6 cases ($\alpha\nabla\alpha$), 3/6 cases (IG) and 3/6 cases (DL) with D-Ref-Reg, compared to 3/6 ($\alpha\nabla\alpha$), 1/6 (IG)  and none (DL) with Pre-def. 
D-Ref-Tok-mask exhibits improvements on average ($\alpha \nabla \alpha$: 57.8, IG: 58.2, DL: 57.8)  over Van-FT (57.3), demonstrating the effectiveness of the token extraction mechanism of D-Ref. Finally, D-Ref-Comb displays the best overall performance, with the highest average score of 59. We attribute this improvement from D-Ref to its  increased coverage with dynamic token extraction, and reduction of spurious source-specific correlations, while the baselines only penalize the group identifiers. A dynamic approach also corrects the model during training before it can get fully biased towards these tokens. Finally, it can incorporate the pre-defined lists along with the extracted tokens, and further improve the performance.

\subsection{Domain-Adaptation Approaches}
We further compare D-Ref-Reg with various Domain Adaptation (DA) methods. However, such methods typically leverage the unlabeled train set from the target domain ($D_T^{train}$). 
We first continue pre-training BERT model on $D_T^{train}$ following \citet{rietzler-etal-2020-adapt}. Then, we perform supervised fine-tuning and regularization on $D_S^{train}$ using D-Ref-Reg (\textbf{M}asked \textbf{L}anguage \textbf{M}odel + D-Ref-Reg). 
We compare against the following methods: 
\paragraph{(i) BERT Van-MLM-FT}: MLM training of BERT on $D_T^{train}$ and supervised fine-tuning on $D_S^{train}$.
\paragraph{(ii) BERT PERL} (Pivot-based Encoder Representation of Language) \citep{ben-david-etal-2020-perl}: This performs pivot based fine-tuning using the MLM objective of BERT by masking and predicting the pivot terms present in the combination of $D_S^{train}$ and the unlabeled $D_T^{train}$. Here pivots are terms that are frequently present in the unlabeled data of both the source and target corpora, and are predictive of the source labels. 
\begin{table*}[!t]
\scriptsize
\centering
\begin{tabularx}{\textwidth}{p{3.8cm}|p{1.3cm}|p{1.3cm}|p{1.3cm}|p{1.3cm}|p{1.3cm}|p{1.3cm}|c}
\toprule
\textbf{Approaches} & \textbf{H \textrightarrow D} & \textbf{D \textrightarrow H} & \textbf{H \textrightarrow W} & \textbf{W \textrightarrow H} & \textbf{D \textrightarrow W}&\textbf{W \textrightarrow D} & \multicolumn{1}{c}{\textbf{Average}}  \\
\hline
BERT Van-MLM-FT & 56.6$\pm$1.3& 66.2$\pm$1.2& 70.0$\pm$2.5&50.9$\pm$2.1& 61.4$\pm$2.4&43.5$\pm$1.9& 58.1\\
BERT PERL& 54.1$\pm$0.7 &60.0$\pm$0.6 &60.1$\pm$2.0 &\textbf{55.2}$\mbox{*}\pm$0.7 &55.5$\pm$1.0 &37.8$\pm$1.2 & 53.8 \\

BERT-AAD& 56.6$\pm$1.3 & 53.9$\pm$3.5 & 68.8$\pm$2.5 &50.7$\pm$1.4& 48.3$\pm$4.7 &\textbf{53.0}$\mbox{*}\pm$1.7& 55.2\\

HATN & 48.4$\pm$1.6 & 59.1$\pm$0.4 & 59.7$\pm$2.9 & 51.4$\pm$1.8 & 60.0$\pm$2.6 & 45.4$\pm$2.7 & 54.0\\
MLM + \citet{Sarwar2021UnsupervisedDA} & 55.0$\pm$1.9 & 66.2$\pm$2.0 & 68.8$\pm$1.1 & 48.2$\pm$3.1 & 57.9$\pm$1.3 & 36.2$\pm$1.1  & 55.4\\
MLM + $\chi^2$-test & 57.9$\pm$1.6 & 67.1$\pm$1.7 & 69.8$\pm$0.8 & 48.2$\pm$3.1 & 60.4$\pm$2.8 & 44.1$\pm$3.4 & 57.9 \\\hline
MLM + D-Ref-Reg ($\alpha \nabla \alpha$) & 57.6$\pm$1.9 & 66.2$\pm$1.2 & \textbf{70.7}$\pm$1.2&52.5$\mbox{*}\pm$4.0 & 62.8$\pm$1.4&48.0$\mbox{*}\pm$4.3 &59.6\\
MLM + D-Ref-Reg (IG) & 58.6$\mbox{*}\pm$1.2 & \textbf{66.8}$\pm$0.5 & 70.1$\pm$1.5 &52.1$\pm$3.0 &62.5$\pm$3.0 &48.9$\mbox{*}\pm$4.4&59.8\\
MLM + D-Ref-Reg (DL) & \textbf{58.8}$\mbox{*}\pm$2.2 & 66.7$\pm$0.6 & 70.5$\pm$1.3 & 
52.4$\mbox{*}\pm$3.5& \textbf{64.7}$\mbox{*}\pm$2.1 & 51.5$\mbox{*}\pm$4.9 & \textbf{60.8}\\
\bottomrule
\end{tabularx}
\caption{\label{Results_DA}
Comparison of DA approaches with D-Ref + MLM.
Macro-F1 ($\pm$std-dev) on different source \textrightarrow target pairs. H : HatEval, D : Dynamic, W : Waseem. $\mbox{*}$ denotes the significantly improved scores w.r.t. Van-MLM-FT.}
\end{table*}

\paragraph{(iii) BERT-AAD} (Adversarial Adaptation with Distillation) \citep{Ryu2020KnowledgeDF}, This is a domain adversarial approach with BERT where a target encoder is adapted with an adversarial objective that leverages $D_S^{train}$ and $D_T^{train}$. 
\paragraph{(iv) HATN} (Hierarchical Attention Transfer Network) \citep{li2018hatn, li2017end} This approach uses attention and a domain adversarial 
pivot extraction mechanism. 
\paragraph{(v)}  \citet{Sarwar2021UnsupervisedDA}:
This  adopts a data-augmentation strategy 
leveraging a negative emotion dataset \citep{Go_Bhayani_Huang_2009}, for cross-domain hate-speech detection. 
They construct a weakly labeled augmented dataset by training a sequence tagger on  $D_S^{train}$ and a TF-IDF based template matching with $D_T^{train}$. 
\paragraph{(vi) $\bm{\chi^2}$-test} using $D_S^{train}$ and $D_T^{train}$.

For a fair comparison, 
we initialize (v) and (vi) with the MLM trained BERT on $D_T^{train}$, while the other methods already make use of $D_T^{train}$ for adaptation.
We use $D_T^{val}$ from target for model selection for all the above methods.

\crefformat{subsection}{\S#2#1#3}

Table \ref{Results_DA} shows the results on comparing against other DA approaches. We note that the average performance of all the other DA approaches in this task is lower than Van-MLM-FT, as discussed in our previous work \citep{bose-etal-2021-unsupervised}. $\chi^2$-test, on an average, fails to surpass the Vanilla baseline. Besides, the DA approach proposed for cross-domain hate-speech detection by \citet{Sarwar2021UnsupervisedDA} also yields an overall drop in performance.
They perform data-augmentation by replacing relevant words from an external negative emotion dataset with tagged hateful terms from the target domain.
We find that a major portion of the augmented instances lack meaning, and this negatively impacts the adaptation.
However, across all feature attribution methods, D-Ref-Reg improves the cross-corpora performance compared to Van-MLM-FT and the DA approaches, with average macro-F1 of 59.6 ($\alpha\nabla\alpha$), 59.8 (IG), and 60.8 (DL), compared to 58.1 from Van-MLM-FT. Since D-Ref-Reg and Van-MLM-FT use identical MLM pre-training on $D_T^{train}$, the improvements can be attributed to the 
dynamic token extraction of our method.

More generally, when the larger set of target domain unannotated instances $D_T^{train}$ are unavailable, D-Ref can identify and correct spurious correlations on source using a small amount of annotated instances from the target $D_T^{val}$, as demonstrated in Section \ref{results_cross}. When sufficient number of unannotated instances from the target corpus are available, D-Ref can yield further cross-corpora improvements by leveraging the unannotated target instances with the MLM pre-training.

\subsection{Qualitative Analysis}
\label{Expl_vis}

\begin{table}[!t]
\scriptsize
\centering
\begin{tabularx}{\columnwidth}{p{3.5cm} p{3.5cm}}
\hline \multicolumn{2}{p{7cm}}{\textbf{Non-hate examples  
\textit{HatEval} \textrightarrow \textit{Dynamic}}}\\
\hline \textbf{FP with Van-FT}& \textbf{TN with D-Ref-Reg (IG)} \\ 
\colorbox{purple!12.0}{Wildlife} \colorbox{purple!67.6}{belongs} \colorbox{purple!29.0}{in} \colorbox{purple!3.6}{the} \colorbox{purple!76.5}{f*cking} \colorbox{purple!22.9}{wild}.  & \colorbox{purple!12.4}{Wildlife} \colorbox{purple!8.3}{belongs} \colorbox{purple!20.0}{in} \colorbox{purple!15.6}{the} \colorbox{purple!7.3}{f*cking} \colorbox{purple!8.0}{wild.}\\\hline
\colorbox{purple!12.870000000000001}{NHS} \colorbox{purple!34.510000000000005}{s*cks} & \colorbox{purple!34.510000000000005}{NHS} \colorbox{purple!12.870000000000001}{s*cks} \\
\hline\hline
\multicolumn{2}{p{7cm}}{\textbf{Hate examples 
\textit{Waseem} \textrightarrow \textit{Dynamic}}}\\
\hline
\textbf{FN with Van-FT}& \textbf{TP with D-Ref-Reg (IG)} \\ 
\colorbox{purple!52.55}{Blacks} \colorbox{purple!10.32}{are} \colorbox{purple!8.41}{savages}. & \colorbox{purple!8}  {Blacks} \colorbox{purple!27.0}{are} \colorbox{purple!18.8}{savages}.\\\hline
\colorbox{purple!90.0}{Queers} \colorbox{purple!1.5611}{can't} \colorbox{purple!2.1342}{be} \colorbox{purple!5.3546}{truly} \colorbox{purple!10.0}{happy}.&\colorbox{purple!8.5}{Queers} \colorbox{purple!25.4}{can't} \colorbox{purple!50.5}{be} \colorbox{purple!36.7}{truly} \colorbox{purple!25.5}{happy}.\\
\hline
\end{tabularx}
\caption{\label{qualit-analysis} Change in attributions with D-Ref-Reg (IG).} 
\end{table}
Table \ref{qualit-analysis} shows the change in attributions for some instances in $D_T^{test}$ from \textit{Dynamic} that were mis-classified by Van-FT but correctly classified by our D-Ref-Reg (IG). 
Van-FT wrongly attributes higher importance to `f*cking' and `s*cks' for the hate class in the first example, and `blacks' and `queers' for non-hate in the second due to source-specific correlations. However, D-Ref-Reg (IG), extracts and penalizes abusive tokens like \{s*ck, a**hole, d*ck\} for the former causing FP and \{africans, dark, queer\} for the latter causing FN.
Our approach not only penalizes the exact tokens, but also those with similar meaning (e.g. `blacks' is contextually close to `dark', `africans'), giving more importance to the context around the spurious tokens.
See Appendix \ref{sec:extraction} for the token-lists.

\section{Conclusion}

We proposed a dynamic approach for automatic token extraction with regularization of the source model such that the spurious source specific correlations are reduced. 
Our approach shows consistent cross-corpora performance improvements both independently and in combination with pre-defined tokens. 
Future work includes applying our method on other 
cross-domain text classification tasks and exploring how explanation faithfulness can be improved in out-of-domain settings~\cite{chrysostomou2022empirical}.

\section*{Ethical Considerations}

The approach proposed in the paper is aimed at supporting robust and accurate detection of online hate speech. The datasets used in the work are publicly available and referenced appropriately. The dataset creators have presented, in detail, the data collection process and annotation guidelines in peer-reviewed articles. The offensive terms presented, as examples, are only intended for better analysis of the models for research purposes.

\section*{Acknowledgements}

This work was supported partly by the french PIA project ``Lorraine Université d'Excellence'', reference ANR-15-IDEX-04-LUE. Experiments presented in this article were carried out using the
Grid'5000 testbed, supported by a scientific interest group hosted by Inria and including CNRS, RENATER and several Universities as well as other organizations (see \url{https://www.grid5000.fr}). We thank the anonymous reviewers for their valuable feedback and suggestions. We would also like to thank George Chrysostomou for his help and suggestions regarding the work during informal discussions.



\bibliographystyle{acl_natbib}
\bibliography{acl2021}






\appendix

\section{Data Description}
\label{data_desc}
While \textit{HatEval} and \textit{Waseem} are sampled from Twitter,
\textit{Dynamic} is generated using a human-and-model-in-the-loop process. These corpora have been collected across different time frames, and hence they involve different topics of discussion, which are also determined to a large extent by the keywords used for sampling. As such, the problem of dataset bias with spurious correlations are induced with such focused sampling procedures \cite{wiegand-etal-2019-detection} used in \textit{Waseem} and \textit{HatEval}. For instance, in \textit{Waseem}, a large amount of tweets, available at the time of our experiments, consist of hate tweets directed against women, which results in False Positives for instances from other corpora that contain women related terms. We observed that most of the racist tweets were already removed and were unavailable for experiments. \textit{HatEval}, on the other hand, has a mix of tweets directed against women and immigrants, and hence it demonstrates decent performance when evaluated over \textit{Waseem} that consists of sexist tweets. On the contrary, \textit{Dynamic} contains annotator-generated tweets that includes challenging perturbations. For instance, it includes non-hate instances like `It’s wonderful having gay people around here', `I hate the concept of hate', `Tea is f*cking disgusting', which can easily fool a classifier learned on biased datasets, and result in classifying these instances as hateful. Moreover, this corpus covers different targets of hate. As such, when \textit{Dynamic} is used as the target corpus, the spurious correlations learned by the source classifier become relatively well-visible, which are captured and penalized by D-Ref while the source model gets trained.

The data used in the work are publicly available, and download links are provided in the respective original articles, which are referenced in this paper. However, in the case of \textit{Waseem}, where only tweet IDs are provided, some tweets might be unavailable.

\section{Implementation Details}
\label{hyper-param}
We leverage the pretrained BERT-base model\footnote{\url{https://github.com/huggingface/transformers}} for our experiments. We  use a batch size of 8, learning rate of  $1 \times 10^{-5}$ and Adam optimizer with decoupled weight decay regularization \citep{Loshchilov2019DecoupledWD} for Van-FT, Van-MLM-FT, D-Ref and Pre-def. For Integrated Gradients, following \citet{liu-avci-2019-incorporating}, the interpolated embeddings are treated as constants while back-propagating the loss from the regularization term. An all zero embedding vector is used as the baseline input for both Integrated Gradients and DeepLIFT.
We use the original code, as provided by the respective authors, for all the prior-arts. For Pre-Def, we combined the pre-defined lists from \citet{kennedy-etal-2020-contextualizing} and \citet{liu-avci-2019-incorporating} and regularized their attribution scores over BERT with $\alpha\nabla\alpha$, IG, and DL as feature attribution methods.

We implement the data-augmentation approach proposed by \citet{Sarwar2021UnsupervisedDA} ourselves due to the absence of an available implementation. Following the description present in the paper, we prepare the training data for the sequence tagger by labeling all the terms in the hateful instances from the source corpus that are also present in the lexicon from hatebase.org\footnote{\url{https://hatebase/org/}}. However, we do not tokenize the lexicon obtained from hatebase.org while searching for the corresponding matching terms in the source corpus. We convert the lexicon into lower-case and look for the exact match in the source corpus.

For D-Ref, we set the value of top $N$ tokens used from ranked \{glist-hate, glist-nhate\} as 500. The values of $k$ $\in$ top \{10\%, 20\%, 30\%, 40\%\} of the instance-length in D-Ref, and $\lambda$ in both D-Ref and Pre-def are selected through hyper-parameter tuning over $D_T^{val}$ using a random seed. For $\alpha \nabla \alpha$ and DeepLIFT, $\lambda$ $\in$ \{0.1, 0.5, 1, 10, 20, 30, 40, 50, 60\} and for IG, $\lambda$ $\in$ \{1, 10, 20, 30, 40, 50, 60\}.  We run supervised fine-tuning on $D_S^{train}$ for 6 epochs with all the BERT models (prior-arts and D-Ref). We select the models (prior-arts and D-Ref) by tuning over $D_T^{val}$ from the target corpus, with respect to macro-F1 scores. Table \ref{Dev-Results_Dyn-Ref} presents the macro-F1 scores obtained on the validation set for D-Ref and the prior arts. 

\begin{table*}[!t]
\small
\centering
\begin{tabularx}{\textwidth}{p{0.6cm}|p{2.5cm}|p{1.4cm}|p{1.4cm}|p{1.4cm}|p{1.4cm}|p{1.4cm}|p{1.4cm}|c}
\toprule
\multicolumn{2}{l|}{\textbf{Approaches}} & \textbf{H \textrightarrow D} & \textbf{D \textrightarrow H} & \textbf{H \textrightarrow W} & \textbf{W \textrightarrow H} & \textbf{D \textrightarrow W}&\textbf{W \textrightarrow D} & \textbf{Average}  \\
\hline
\multicolumn{2}{l|}{BERT Van-FT} & 54.7$\pm$0.8 & 64.7$\pm$1.1&65.6$\pm$4.5 & 59.4$\pm$1.2 & 61.9$\pm$1.1 & 46.9$\pm$4.7 & 58.9\\
\multicolumn{2}{l|}{\citet{liu-avci-2019-incorporating}} & 45.3$\pm$5.2 & 50.3$\pm$1.1 & 57.1$\pm$2.4 & 49.7$\pm$0.5 & 56.8$\pm$3.2 &39.3$\pm$2.0 & 49.8\\
\multicolumn{2}{l|}{\citet{kennedy-etal-2020-contextualizing} (a)} &53.5$\pm$1.1 &62.8$\pm$1.5& 60.3$\pm$2.5 & 53.9$\pm$8.8 &51.3$\pm$2.3& 43.6$\pm$2.3& 54.2\\
\multicolumn{2}{l|}{\citet{kennedy-etal-2020-contextualizing} (b)} & 54.8$\pm$4.2 & 55.5$\pm$3.9 & 62.1$\pm$1.8& 61.3$\pm$0.9 & 58.6$\pm$4.4 & 46.3$\pm$2.7 & 56.4 \\
\hline
\hline
\multicolumn{2}{l|}{Pre-def ($\alpha \nabla \alpha$)} & 55.2$\pm$1.0 & 65.8$\pm$1.1 & 67.8$\pm$3.4 & 59.2$\pm$1.0 & 62.1$\pm$1.9 &47.2$\pm$4.1 & 59.6\\
\multicolumn{2}{l|}{D-Ref-Tok-rem ($\alpha \nabla \alpha$)} & 54.9$\pm$1.0 & 64.7$\pm$1.2 &66.6$\pm$2.8  & 58.5$\pm$1.4 & 60.7$\pm$1.1& 45.9$\pm$3.9 & 58.6\\
\multicolumn{2}{l|}{D-Ref-Reg ($\alpha \nabla \alpha$)} & 55.4$\pm$0.7 & 65.4$\pm$1.9 & 65.5$\pm$3.9 & 59.5$\pm$0.9 &  61.0$\pm$1.1& 49.6$\pm$3.7 & 59.4 \\
\multicolumn{2}{l|}{D-Ref-Comb ($\alpha \nabla \alpha$)} &  56.2$\pm$1.7& 64.6$\pm$0.7 & 66.8$\pm$2.9 & 59.9$\pm$1.3 & 62.6$\pm$1.7 & 48.1$\pm$1.1 & 59.7\\
\hline
\hline
\multicolumn{2}{l|}{Pre-def (IG)} &  55.7$\pm$1.6 & 64.8$\pm$0.7 &67.0$\pm$2.2  & 59.9$\pm$0.9 & 62.3$\pm$1.7 & 44.4$\pm$3.2 & 59.0\\
\multicolumn{2}{l|}{D-Ref-Tok-rem (IG)} & 56.5$\pm$1.9 & 63.5$\pm$1.4 & 65.4$\pm$2.0 & 59.0$\pm$1.1 & 59.9$\pm$0.8& 49.7$\pm$1.9 & 59.0\\
\multicolumn{2}{l|}{D-Ref-Reg (IG)} & 57.5$\pm$2.1 & 64.8$\pm$1.3 & 67.1$\pm$2.3 & 59.6$\pm$1.3& 60.3$\pm$1.1 &47.7$\pm$4.0 & 59.5\\
\multicolumn{2}{l|}{D-Ref-Comb (IG)}& 57.2$\pm$0.8 & 64.3$\pm$1.5 & 67.4$\pm$2.5 &  58.3$\pm$0.9& 62.0$\pm$1.5 & 52.1$\pm$3.7 & 60.2\\
\hline \hline
\multicolumn{2}{l|}{Pre-def (DL)} & 54.5$\pm$2.1 & 65.1$\pm$1.1 & 66.1$\pm$1.4 & 60.1$\pm$0.4 & 61.3$\pm$1.4&45.1$\pm$1.7& 58.7\\
\multicolumn{2}{l|}{D-Ref-Tok-rem (DL)} & 55.4$\pm$1.9 & 65.5$\pm$1.5 & 65.5$\pm$3.1 & 59.0$\pm$1.1 &61.6$\pm$2.1& 48.3$\pm$3.7 & 59.2\\
\multicolumn{2}{l|}{D-Ref-Reg (DL)} & 56.0$\pm$1.8 & 65.7$\pm$0.8 & 68.1$\pm$2.3 & 59.3$\pm$1.4 & 63.0$\pm$1.4  & 48.0$\pm$5.9& 60.0\\
\multicolumn{2}{l|}{D-Ref-Comb (DL)} & 55.1$\pm$2.1 & 65.6$\pm$1.4 &66.4$\pm$3.1  &59.1$\pm$0.8  & 61.6$\pm$2.2 & 49.6$\pm$3.0& 59.6 \\
\hline
\end{tabularx}
\caption{\label{Dev-Results_Dyn-Ref}
Validation set ($D_T^{val})$ macro F1 ($\pm$std-dev) on source \textrightarrow target pairs (H : HatEval, D : Dynamic, W : Waseem).}
\end{table*}

\section{Tokens extracted in different epochs}
\label{sec:extraction}

The list of error-causing tokens for False Positives (FP) and False Negatives (FN) in $D_T^{val}$, extracted for the cases presented in Section \ref{Expl_vis}, is given below. We underline the tokens present in the visualization examples (both Table \ref{qualit-analysis} in Section \ref{Expl_vis} and below) and ones similar in meaning to them.
\newline
\textbf{HatEval \textrightarrow Dynamic}
\begin{itemize}
    \item \textbf{Epoch 1:}
\textbf{FP:} \{idiots, conservative, countries, p*ssy, bloody, americans, move, \underline{a**hole}, hating, beings, feminist, africans, resources, \underline{d*ck}, resist, \underline{females}, attacks, dude, anger \}
\textbf{FN:} \{hitler, plague, \#\#urs, crisis, rescue, funding, gorgeous, treason, journalist, lawyers, agenda, roles, principles, bloody, intern\}
    \item \textbf{Epoch 2:}
\textbf{FP:} \{race, hating, flights, sheep, \underline{females}, ignorant, feminist, resist, attacks, \underline{d*ck}, kill, boat, countries, p*ssy, refugee, bloody\}
\textbf{FN:} \{president, foreigners, illegal, betrayal, lgbt, riots, gorgeous, treason, joking, chris, intelligent, arguments, humans\}
    \item \textbf{Epoch 3:}
\textbf{FP:} \{countries, race, hating, \underline{females}, feminist, africans, ridiculous, \underline{d*ck}, express, comments, organized, \underline{s*ck}, allow, bloody\}
\textbf{FN:} \{illegal, hitler, generally, david, intelligent, secret, chris, equality, dating, yellow, treason, abuses, \#\#gb, humans, plague, dear, nonsense\}
    \item \textbf{Epoch 4:}
\textbf{FP:} \{isis, genocide, indians, society, supported, \underline{females}, feminist, attacks, \underline{s*ck}, destroy, migrants \}
\textbf{FN:} \{hitler, opportunities, sister, betrayal, \#\#ame, gorgeous, \#\#heads, dating, riots, bank, murders,
arguments, humans, fights, plague, influence, targeting, supporters, coordination, lies, \#\#boys\}
    \item \textbf{Epoch 5:}
\textbf{FP:} \{clean, ignorant, slave, feminist, punish, africans, \#\#ache, \underline{d*ck}, \#\#fs, ars, destroy, status, race, p*ssy, western, send\}
\textbf{FN:} \{statement, gross, hitler, sending, yellow, waste, hopefully, trapped, riots, bait, sister, coordination, humans\}
   \item \textbf{Epoch 6:}
\textbf{FP:} \{soft, suicide, countries, p*ssy, bloody, genocide, punish, destroy, migrants, vile, beings, savage, feminist, tory, awful, ignorant, \#\#ists, spend, send\}
\textbf{FN:} \{gross, secret, influence, yellow, crime, abuses, participate, approach\}
\end{itemize}

\begin{table*}[!htb]
\small
\centering
\begin{tabularx}{\textwidth}{p{3.8cm}|p{1.2cm}|p{1.2cm}|p{1.2cm}|p{1.4cm}|p{1.4cm}|p{1.33cm}|c}
\hline
\multicolumn{1}{l|}{\textbf{Approaches}} & \multicolumn{2}{c|}{\textbf{HatEval}} & \multicolumn{2}{c|}{\textbf{Dynamic}} & \multicolumn{2}{c|}{\textbf{Waseem}} & \textbf{Average}  \\
\hline
BERT Van-FT & \multicolumn{2}{c|}{43.3$\pm$1.8} & \multicolumn{2}{c|}{85.1$\pm$0.5} & \multicolumn{2}{c|}{85.4$\pm$0.7} & 71.3 \\
\hline
\multicolumn{8}{c}{\textbf{In-corpus performance on source (left of arrows) while refining the source model for the target (right of arrows)}}\\ \hline
 & \textbf{H \textrightarrow D} & \textbf{H \textrightarrow W} & \textbf{D \textrightarrow H}  & \textbf{D \textrightarrow W} & \textbf{W \textrightarrow H} &\textbf{W \textrightarrow D} &   \\
\hline
D-Ref-Reg ($\alpha \nabla \alpha$) & 39.7$\pm$3.2 &38.4$\pm$1.7 &84.1$\pm$1.0  & 84.2$\pm$0.8 &84.4$\pm$0.7 &78.8$\pm$8.0&68.3\\
D-Ref-Reg (IG) & 40.5$\pm$2.0 &37.7$\pm$2.1 &84.0$\pm$0.4 &84.5$\pm$0.4 & 84.6$\pm$1.0 &85.3$\pm$1.4&69.4\\
D-Ref-Reg (DL) & 37.1$\pm$1.8 & 38.1$\pm$2.9 & 84.7$\pm$0.6 &84.3$\pm$1.2& 84.4$\pm$0.5 & 80.7$\pm$6.4& 68.2\\
\hline
\end{tabularx}
\caption{\label{In-corpus}
In-corpus macro F1 ($\pm$std-dev), i.e. the source corpus performance, obtained after refining the source model for the target corpus (present at the right hand side of the arrows) using D-Ref-Reg. H : HatEval, D : Dynamic, W : Waseem. For D-Ref-Reg, model-selection and early-stopping is done over the validation set from the target corpus.}
\end{table*}

A non-hate comment in \textit{Dynamic} test set for the above case, wrongly classified as hate by Van-FT and correctly classified as non-hate with D-Ref-Reg (IG), is given below. Darker the shade, higher is the attribution:

\begin{CJK*}{UTF8}{gbsn}
{\setlength{\fboxsep}{0pt}\colorbox{white!0}{\parbox{0.9\columnwidth}{
\textbf{Van-FT}: \colorbox{purple!0.61109}{\strut There} \colorbox{purple!0.28684}{\strut is} \colorbox{purple!0.16897}{\strut so} \colorbox{purple!6.3843000000000005}{\strut much} \colorbox{purple!20.0}{\strut cancer} \colorbox{purple!9.5107}{\strut patients} \colorbox{purple!6.029}{\strut in} \colorbox{purple!10.0}{\strut the} \colorbox{purple!20.0}{\strut world} \colorbox{purple!2.7869}{\strut but} \colorbox{purple!0.23001}{\strut it} \colorbox{purple!0.28684}{\strut is} \colorbox{purple!20.0}{\strut mostly} \colorbox{purple!10.0}{\strut the} \colorbox{purple!29.999999999999996}{\strut young} \colorbox{purple!59.99999999999999}{\strut females} \colorbox{purple!10.0}{\strut who} \colorbox{purple!2.8157}{\strut are} \colorbox{purple!10.0}{\strut worstly}
\colorbox{purple!20.0}{\strut affected} \colorbox{purple!10.0}{\strut by} \colorbox{purple!29.999999999999996}{\strut this} \colorbox{purple!29.999999999999996}{\strut disease.}
}}}

\end{CJK*}
\begin{CJK*}{UTF8}{gbsn}
{\setlength{\fboxsep}{0pt}\colorbox{white!0}{\parbox{0.9\columnwidth}{
\textbf{D-Ref-Reg}: \colorbox{purple!12.3}{\strut There} \colorbox{purple!25.9}{\strut is} \colorbox{purple!15.299999999999999}{\strut so} \colorbox{purple!3.5}{\strut much} \colorbox{purple!41.6}{\strut cancer} \colorbox{purple!12.1}{\strut patients} \colorbox{purple!7.0}{\strut in} \colorbox{purple!15.7}{\strut the} \colorbox{purple!4.1000000000000005}{\strut world} \colorbox{purple!2.9}{\strut but} \colorbox{purple!6.4}{\strut it} \colorbox{purple!15.299999999999999}{\strut is} \colorbox{purple!2.7}{\strut mostly} \colorbox{purple!0.8}{\strut the} \colorbox{purple!2.9}{\strut young} \colorbox{purple!24.1}{\strut females} \colorbox{purple!6.1000000000000005}{\strut who} \colorbox{purple!14.5}{\strut are} \colorbox{purple!66.8}{\strut worstly} \colorbox{purple!29.7}{\strut affected} \colorbox{purple!16.400000000000002}{\strut by} \colorbox{purple!12.7}{\strut this} \colorbox{purple!18.4}{\strut disease.}
}}}
\end{CJK*}
\newline
\newline
\textbf{Waseem \textrightarrow Dynamic}
\begin{itemize}
    \item \textbf{Epoch 1:}
\textbf{FP:} \{female, \#\#ists, fe, sex, feminist, rap\}
\textbf{FN:} \{cast, coward, queer, equality, \#\#bi , cost, \#\#sy, born, \underline{asian}, nazis, kids, cancer, gender, hiring, funded\}
    \item \textbf{Epoch 2:}
\textbf{FP:} \{\#\#ists, her, sex, worse, feminist, \#\#nt, outraged\}
\textbf{FN:} \{welcome, caused, cancer. drag, \#\#bi, pressure, parent, nazis, troll, cast, trash, ruins, lesbian, attacking, chinese\}
    \item \textbf{Epoch 3:}
\textbf{FP:} \{female, \#\#ja, might, men, feminist\}
\textbf{FN:} \{quoting, govt, referring, nazis, troll, lesbian, rogue, date, chinese, typically\}
    \item \textbf{Epoch 4:}
\textbf{FP:} \{communism, her, openly, intelligent, many, barbie, chicks, females, arguing\}
\textbf{FN:} \{date, suggest, \#\#lat, referring, police, chinese, cancer, voice, native, lesbian\}
   \item \textbf{Epoch 5:}
\textbf{FP:} \{term, f*ck, \#\#ng, woman, \#\#ist, feminist, females, prison\}
\textbf{FN:} \{removed, educate, freaking, \underline{queer}, wow, ending, referring, dye, \#\#wat, issues, \underline{africans}, vast, chinese, \underline{dark}\}
   \item \textbf{Epoch 6:}
\textbf{FP:} \{whore, her, \#\#ots, role, sweden, pay, d*ck, trump, feminist, females, american, arguing\}
\textbf{FN:} \{bat, everyday, freak, argument, movement, chinese, tho, feature, lesbian\}
\end{itemize}

A hate comment in \textit{Dynamic} test set for the above case, wrongly classified as non-hate by Van-FT and correctly classified as hate with D-Ref-Reg (IG), is given below. Darker the shade, higher is the attribution:

\begin{CJK*}{UTF8}{gbsn}
{\textbf{Van-FT}: \setlength{\fboxsep}{0pt}\colorbox{white!0}{\parbox{0.7\columnwidth}{
\colorbox{purple!20.4}{\strut Don't} \colorbox{purple!0.6}{\strut get} \colorbox{purple!9.6}{\strut me} \colorbox{purple!18.9}{\strut wrong} \colorbox{purple!6.0}{\strut I} \colorbox{purple!14.100000000000001}{\strut don't} \colorbox{purple!24.6}{\strut hate} \colorbox{purple!54.9}{\strut asians,} \colorbox{purple!2.4}{\strut but} \colorbox{purple!12.0}{\strut I} \colorbox{purple!13.799999999999999}{\strut definitely} \colorbox{purple!10.799999999999999}{\strut don't} \colorbox{purple!16.5}{\strut like} \colorbox{purple!15.3}{\strut them}
}}}
\end{CJK*}

\begin{CJK*}{UTF8}{gbsn}
{\textbf{D-Ref-Reg}: \setlength{\fboxsep}{0pt}\colorbox{white!0}{\parbox{0.7\columnwidth}{
\colorbox{purple!45.199999999999996}{\strut Don't} \colorbox{purple!46.4}{\strut get} \colorbox{purple!4.3}{\strut me} \colorbox{purple!28.8}{\strut wrong} \colorbox{purple!2.8}{\strut I} \colorbox{purple!14.6}{\strut don't} \colorbox{purple!8.0}{\strut hate} \colorbox{purple!8.6}{\strut asians,} \colorbox{purple!71.6}{\strut but} \colorbox{purple!16.0}{\strut I} \colorbox{purple!13.2}{\strut definitely} \colorbox{purple!9.799999999999999}{\strut don't} \colorbox{purple!11.6}{\strut like} \colorbox{purple!9.2}{\strut them}
}}}
\end{CJK*}

Since, the \textit{Waseem} dataset is made available as tweet IDs, we observed that it mostly contains sexist comments, while most of the racist content must have been removed before we could crawl it. Hence, the tokens related to race mostly occur in non-hate contexts causing FN. 

Even though some error-causing tokens remain in the list until the end, their overall effect should be reduced as the regularization is performed throughout the training procedure, which causes improvement in macro F1.

\section{In-corpus performance}
\label{in-corpus}

We present the in-corpus performance, i.e. the performance on the source corpus in terms of macro-F1 scores, obtained when the source model is refined for the corresponding target corpus using D-Ref-Reg, in Table \ref{In-corpus}. For D-Ref-Reg, the model is tuned over the target corpus validation set. Here BERT Van-FT gives the original performance of the source model, when no refinement is performed, as a reference. In this case, the model is tuned over the in-corpus validation set. The \textit{HatEval} corpus is part of a shared task and involves a challenging test set with low in-corpus performance. 
The drop across in-corpus performance with D-Ref-Reg is expected, as the main goal of the proposed approach is to make the source model best suited for the target corpus.

\section{Pre-defined group identifiers}
\label{pre-def}
The combined list of pre-defined group identifiers from \citet{liu-avci-2019-incorporating} and \citet{kennedy-etal-2020-contextualizing} are given below:

\{lesbian, gay, bisexual, trans, cis, queer, lgbt, lgbtq, straight, heterosexual, male, female, nonbinary, african, african american, european, hispanic, latino, latina, latinx, canadian, american, asian, indian,
middle eastern, chinese, japanese, christian, buddhist, catholic, protestant, sikh, taoist, old, older, young, younger, teenage, millenial, middle aged, elderly, blind, deaf, paralyzed, muslim, jew, jews, white, islam, blacks, muslims, women, whites, gay, black, democrat, islamic, allah, jewish, lesbian, transgender, race, brown, woman, mexican, religion, homosexual, homosexuality, africans\}

\begin{table}[!t]
\centering
\small
\begin{tabularx}{\columnwidth}{p{2.65cm}|p{1.2cm}|p{1.19cm}|p{1.25cm}}
\toprule
{\textbf{Approaches}} & \textbf{HatEval} & \textbf{Dynamic} & \textbf{Waseem} \\
\hline
{BERT Van-FT} & 1 m 25 s & 3 m 52 s & 2 m \\
{D-Ref-Reg ($\alpha\nabla\alpha$)} & 1 m 39 s & 7 m & 3 m 33 s\\
{D-Ref-Reg (IG)} & 9 m 37 s & 59 m & 19 m 7 s \\
{D-Ref-Reg (DL)} & 4 m 4 s & 18 m 36 s & 8 m 44 s \\

\hline
\end{tabularx}
\caption{\label{Computation_time}
Per epoch training time on different source corpora.}
\end{table}

\section{Computational Efficiency}

We present the per epoch training time for D-Ref-Reg with different source corpora in Table \ref{Computation_time}. 
The training times of D-Ref-Reg ($\alpha \nabla \alpha$) are less than 2 times of that with Van-FT. With D-Ref-Reg (DL), the training time is approximately 4.5 times of that with Van-FT. This demonstrates the computational efficiency of our approach. In the case of D-Ref-Reg (IG), the computation time is indeed high. This occurs due to the aggregation of gradients using a path integral and computing gradients over gradients, as also discussed in \citet{kennedy-etal-2020-contextualizing,liu-avci-2019-incorporating}. However, our approach is not dependent on any particular feature attribution method, as demonstrated with our experiments.



\end{document}